%% file: neurips_2026.tex
\documentclass{article}

\usepackage[preprint]{neurips_2026}

\usepackage[utf8]{inputenc} 
\usepackage[T1]{fontenc}    
\usepackage{hyperref}       
\usepackage{url}            
\usepackage{booktabs}       
\usepackage{amsfonts}       
\usepackage{nicefrac}       
\usepackage{microtype}      
\usepackage{xcolor}         
\usepackage{multirow}
\usepackage{graphicx}
\usepackage{amsmath} 

\usepackage{wrapfig}
\usepackage{subcaption} 

\usepackage[most]{tcolorbox}

\newtcolorbox{mybox}{
  colback=gray!10,     
  colframe=black,      
  sharp corners,       
  boxrule=0.5pt,       
  breakable,           
  enhanced,            
  left=5pt, right=5pt, 
}

\tcbuselibrary{breakable, skins}
\newtcolorbox{promptblock}[1][]{
    breakable,                 
    enhanced,                  
    fontupper=\small\sffamily, 
    colback=gray!5!white,      
    colframe=gray!80!black,    
    boxrule=0.8pt,              
    arc=4pt,                    
    left=8pt, right=8pt,       
    top=8pt, bottom=8pt,       
    drop shadow=black!10,       
    title=#1,                   
    fonttitle=\bfseries\small,  
    coltitle=white,             
    attach boxed title to top left={yshift=-2mm, xshift=4mm}, 
    boxed title style={colback=gray!80!black, arc=2pt},       
    #1                       
}

\title{Generating Biomedical Fact-Checking Reports with RL-Enhanced Agentic Search}

\author{
Jiongxiao Wang$^{1}$\quad Dingli Ma$^{2}$ \quad Chaoqun Ni$^{1}$ \\[3pt]
\textsuperscript{1}University of Wisconsin--Madison; 
\textsuperscript{2}University of Washington
}

\newcommand{\name}{BioCheck Agent}

\begin{document}

\maketitle

\begin{abstract}
Automated fact-checking is essential for ensuring the reliability of public health information, yet the biomedical domain poses unique challenges. Validating biomedical claims requires rigorous interpretation of scientific literature, assessment of retrieved evidence, and comprehensive justification toward the conclusion. Although Large Language Models (LLMs) enhanced by Retrieval-Augmented Generation (RAG) and agentic search perform automated fact-checking in a retrieve-then-verify paradigm, current methods still output isolated prediction labels, lacking explanatory depth and offers limited utility for human understanding. To bridge this gap, we introduce an LLM-based agent named \name{} that generates structured biomedical fact-checking reports with agentic search. Rather than merely outputting supported or refuted labels, our agent synthesizes final conclusions with retrieved evidence and rigorous analysis. To ensure domain-specific accuracy, \name{} exclusively searches high-quality scientific literature in PubMed, utilizing advanced Boolean search operators. Recognizing that direct prompting often results in hallucinations and low-quality reports, especially for lightweight open-source models, we further propose the Evidence-Grounded Group Relative Policy Optimization (EG-GRPO) to perform reinforcement learning on \name{} with a task-specific reward that incentivizes advanced search behavior and high-quality evidence retrieval while penalizing hallucinations. Our experimental results show that compared to the base model Qwen3.5-4B, \name{} with EG-GRPO improves label prediction accuracy on SciFact by 9.95\%. Furthermore, it achieves a 3.7\% higher evidence quality score and a 19.63\% lower evidence hallucination rate, demonstrating its ability to generate biomedical fact-checking reports with improved accuracy and quality.
\end{abstract}

\input{1_introduction}
\input{2_related_work}
\input{3_method}
\input{4_experiment}

\input{5_conclusion}

\bibliographystyle{unsrt}
\bibliography{references}

\clearpage


\appendix

\input{appendix}



\end{document}

%% file: 1_introduction.tex
\section{Introduction}
Health misinformation has become a major challenge for public health, especially during health crises when people must navigate false or misleading claims, information overload, uncertainty, and rapidly changing scientific evidence. The World Health Organization described this problem as an "infodemic" and emphasized that managing misinformation is part of controlling public health emergencies\cite{omojunikanbi2022public}. This challenge is especially consequential in biomedical contexts, where inaccurate claims may shape vaccination decisions, treatment choices, adherence to public health guidance, trust in medical institutions, and the use of unsafe remedies. Prior research shows that health misinformation can produce measurable harms. During the COVID-19 pandemic, misinformation about prevention and treatment circulated widely across countries and was linked to serious real-world consequences, including hospitalizations and deaths\cite{islam2020covid}. Experimental studies further show that exposure to COVID-19 vaccine misinformation can reduce vaccination intent\cite{loomba2021measuring}, while cross-national survey research finds that susceptibility to misinformation is associated with lower self-reported compliance with public health guidance and lower willingness to be vaccinated\cite{roozenbeek2020susceptibility}. Together, these findings suggest that biomedical misinformation is not merely a communication problem; it can affect health behavior, undermine public trust, and weaken evidence-based public health interventions.

Automated fact checking offers a promising way to improve the reliability of online health information, but biomedical fact checking remains particularly difficult. Public health fact checking often requires domain expertise and assessment against credible evidence, and prior work has argued that this setting requires not only veracity prediction but also explanation generation \cite{kotonya2020explainable}. Health claims also require careful interpretation of biomedical evidence, including the strength and certainty of available studies, which is why recent medical fact checking datasets incorporate evidence levels in addition to labels such as supported, refuted, and not enough information \cite{vladika2024healthfc}. More broadly, evidence based medical reasoning requires attention to the population or patient problem, intervention, comparison, and outcome when formulating clinical questions and searching for evidence \cite{schardt2007utilization}. For this reason, biomedical fact-checking should provide more than a final label. It should explain what evidence was retrieved, how that evidence relates to the claim, what uncertainty remains, and why a particular conclusion is justified.

\begin{figure*}[ht]
    \centering
    \includegraphics[width=0.95\textwidth]{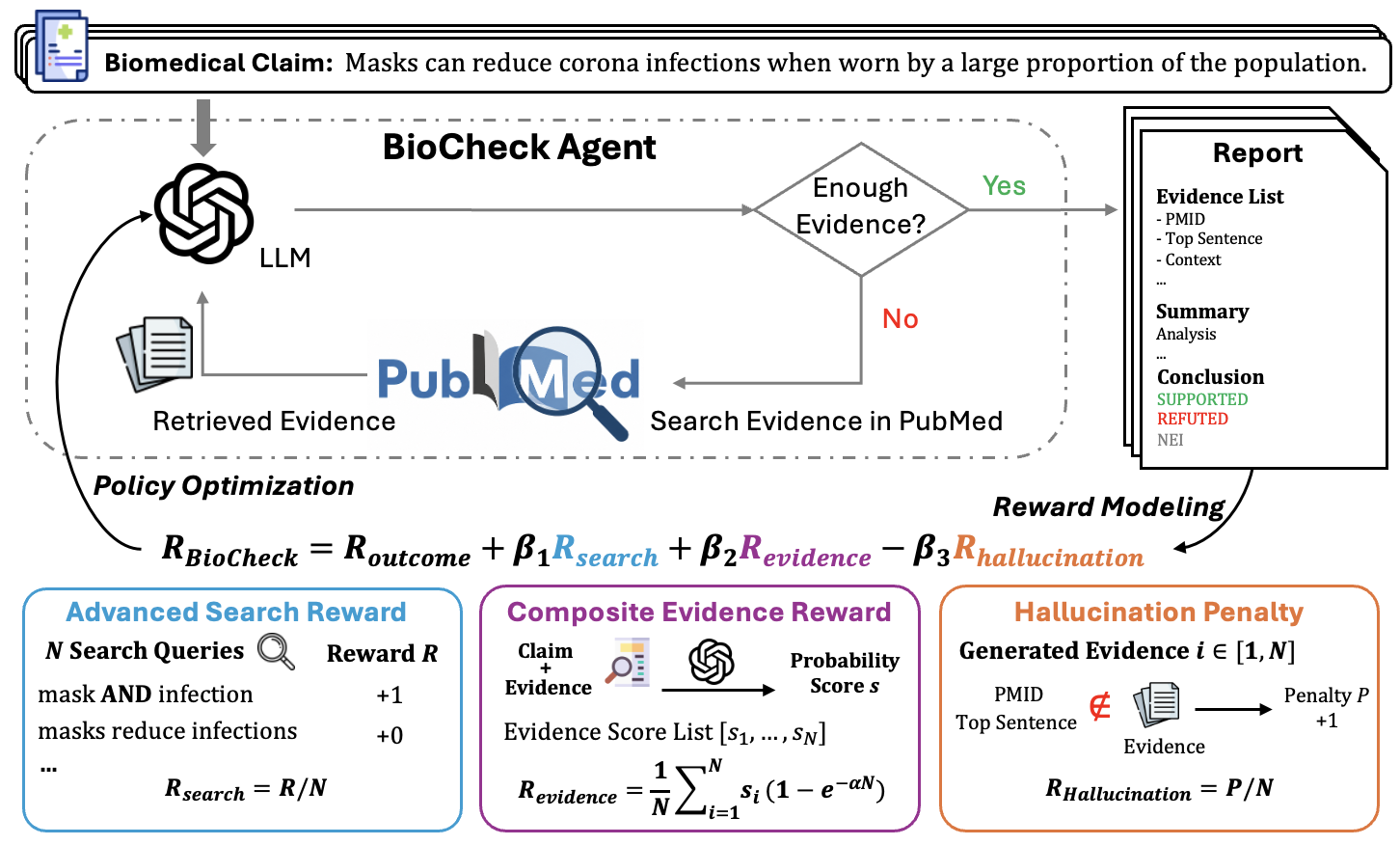}
    \caption{Illustration for \name{} and Reward Modeling for EG-GRPO.}
    \label{fig:intro}
\end{figure*}

With the advancement of Large Language Models (LLMs) and techniques like retrieval-augmented generation (RAG) \cite{lewis2020retrieval} and agentic search \cite{jin2025search}, current automated fact-checking tasks have transitioned from text classification to a retrieve-then-verify paradigm \cite{kim2023factkg, singal2024evidence, wang2024factcheck}. This requires first retrieving external information from the open world and then verifying the claim based on those findings. Although these approaches have achieved high classification accuracy, their label-focused results and evaluation lack perfect reliability, making human experts still essential for robust decision-making where predicted labels fail to provide meaningful information. 

To advance automated biomedical fact-checking beyond label prediction through comprehensive evidence analysis, we propose \textbf{\name{}}, a novel LLM-based agent for structured biomedical fact-checking report generation. \name{} generates comprehensive reports by outlining relevant evidence, writing analytical summaries and making the final conclusion with predicted labels. Tailored for the biomedical domain, our agent executes agentic searches within the PubMed database via its API, which retrieves the most up-to-date literature, thus offering superior evidence quality compared to static databases or generalized search engines. Besides, our agent is equipped with advanced academic search strategies, automatically decomposing complex claims into core entities and linking them with Boolean operators (`AND', `OR') to construct highly accurate search queries.

To further enhance the performance and scalability of \name{} beyond direct prompting, we propose Evidence-Grounded Group Relative Policy Optimization (EG-GRPO) for end-to-end reinforcement learning (RL). Building upon GRPO \cite{shao2024deepseekmath}, which has proven effective across various agentic tasks \cite{feng2025group, zhang2025agentrl}, EG-GRPO employs a task-specific reward design that emphasizes the quality and groundedness of evidence in the generated reports. Specifically, alongside an outcome-based reward for ground-truth label alignment, we introduce three extra components: an Advanced Search Reward to encourage the use of Boolean search operations, a Composite Evidence Reward to balance the quantity and quality of the generated evidence, and a Hallucination Penalty to prevent the model from generating evidence that contradicts the retrieved literature. Then we apply EG-GRPO to train the lightweight Qwen3.5-4B model on the SciFact \cite{wadden2020fact} training split to serve as the backbone model for \name{}. An illustrative overview for \name{} and the reward design of EG-GRPO is shown in Figure~\ref{fig:intro}.

We evaluate the performance of \name{} with EG-GRPO on both the SciFact test split and the HealthFC \cite{vladika2024healthfc} dataset. Across both datasets, \name{} optimized with EG-GRPO outperforms all baselines in label prediction metrics when using Qwen3.5-4B as the base model, even surpassing the commercial model GPT5.2. Compared to the base model prior to RL, EG-GRPO significantly improves prediction accuracy by 9.95\% on SciFact and 7.03\% on HealthFC. Furthermore, although directly applying the baseline GRPO yields performance comparable to EG-GRPO on SciFact, it shows limited improvement on HealthFC. This demonstrates the necessity of our task-specific reward design, where explicitly optimizing for fact-checking reports leads to better transferable performance. Beyond standard label prediction metrics, we introduce two additional metrics designed to provide fine-grained evaluations of the generated reports: Evidence Quality Score (EQS) and Evidence Hallucination Rate (EHR). Compared to the base model Qwen3.5-4B, applying EG-GRPO on \name{} clearly improves the quality of the generated biomedical fact-checking reports. Specifically, on the SciFact dataset, it yields a 3.7\% improvement in EQS (reflecting better evidence retrieval) and a substantial 19.63\% reduction in EHR (indicating fewer hallucinations). These results highlight the effectiveness of EG-GRPO in enabling \name{} to achieve state-of-the-art performance in both accurate label prediction and high-quality report generation. Ultimately, by mitigating hallucinations and grounding outputs in high-quality evidence, our \name{} represents a vital step toward assisting humans in discerning health misinformation through generated fact-checking reports.

%% file: 2_related_work.tex
\section{Related Work}
\textbf{Automated Fact-Checking.}
Fact-checking is defined as the task of assessing the veracity of factual claims. Traditionally, fact-checking is performed by human experts in journalism or scientific domains. However, these processes are normally labor-intensive and time-consuming. With the development of Natural Language Processing (NLP), it has become possible to perform automated fact-checking \cite{zeng2021automated, guo2022survey}. Early automated fact-checking mainly depended on supervised learning \cite{vlachos2014fact, lee2020language, wadden2022multivers} to train classifiers for verification, and the evidence was often sourced from pre-annotations within the dataset \cite{thorne2018fever, wadden2020fact}. The emergence of LLMs and RAG has transitioned fact-checking to an open-world retrieve-and-verify paradigm. Rather than relying on static, pre-annotated evidence, current methods actively retrieve information from external sources like vector-based retrieval databases \cite{singal2024evidence, barone2025combining} or structured knowledge graphs \cite{kim2023kg}. More recently, LLM-based agents have gained prominence by leveraging external tools, such as search engines, to perform more comprehensive and real-time evidence retrieval \cite{wang2024factcheck, wei2024long, xie2025fire}. Specifically, biomedical fact-checking presents unique challenges that demand extensive domain expertise and rigorous verification, given its significant impact on public health, as evidenced during the COVID-19 pandemic \cite{apuke2021fake}. To advance automated fact-checking in this field, numerous datasets and benchmarks have been established \cite{wadden2020fact, sarrouti2021evidence, mohr2022covert, vladika2024healthfc, reisle2025evaluating}, alongside specialized methods designed to leverage high-authority, rigorous resources for verification \cite{liu2024retrieval, barone2025combining}.

\textbf{LLM-based Agents and Agentic Search.}
The advanced instruction-following and reasoning capabilities of LLMs now enable the automated execution of complex tasks as agents. Consequently, various frameworks, such as ReAct \cite{yao2022react}, Plan-and-Act \cite{erdogan2025plan}, and CodeAct \cite{wang2024executable}, have been proposed to build autonomous agents with any backbone LLMs. Given the growing real-world impact of agents, industry-level frameworks like LangGraph \cite{langgraph} and OpenHands \cite{wang2024openhands} have also emerged. Besides, most LLMs nowadays are trained to execute predefined actions via a standard function-calling pipeline \cite{function_calling}, rather than relying on structured text outputs like ReAct. As a critical application of LLM-based agents, agentic search integrates search engines with reasoning capabilities to enhance response quality \cite{jin2025search, li2025search}. By grounding outputs in verifiable references, this approach significantly mitigates hallucinations \cite{li2025mitigating}. Initially pioneered by agents such as Bing Chat \cite{kelly2023bing}, agentic search is now widely employed to bolster performance across a broad spectrum of applications, from general benchmarks to domain-specific tasks \cite{deep_research}.

\textbf{Reinforcement Learning for Agents.}
Unlike general chatbot LLMs, whose performance mainly depends on human preferences \cite{ouyang2022training}, agent tasks often feature explicit answers, making reinforcement learning with verifiable rewards (RLVR) highly applicable. Specifically, Group Relative Policy Optimization (GRPO) \cite{shao2024deepseekmath} has been demonstrated as an effective algorithm and is widely applied to improve agent performance across various tasks \cite{jin2025search,wei2025webagent,wang2025reinforcement,feng2025video}.

%% file: 3_method.tex
\section{Method}
This section provides a detailed description of our \name{} and the EG-GRPO method used to further improve the agent performance.

\subsection{\name{}}
\name{} is built as a standard search agent, iteratively invoking an LLM through LangChain to first reason and then call a predefined search tool to retrieve related information for verifying a given claim. Following each tool response, the LLM determines whether to continue searching or to generate a structured fact-checking report using the gathered evidence. To prevent the context window from exceeding its limits, we impose a hard constraint on the agent: once a maximum number of searches is reached, we inject a user instruction as the intervention. This prompts the LLM to terminate the search and immediately output its report based on the evidence collected thus far. 

\textbf{Agent Framework.} We utilize LangGraph \cite{langgraph} to develop \name{}. LangGraph is a low-level orchestration framework designed for building stateful agents via graph-based structures. Within this framework, each node represents an action, such as invoking LLMs, executing tools, or incorporating human-in-the-loop interventions, while edges define the execution flow between nodes, often governed by conditional logic. Additionally, since LangGraph only serves as a framework to connect agent nodes, LangChain \cite{langchain} remains essential for defining key nodes by invoking LLMs with customized tools through their standard function-calling capabilities.

\textbf{Tool Definition.} Given our focus on biomedical fact-checking, we designed the search tool named $pubmed\_search$ that utilizes the PubMed API \footnote{https://www.ncbi.nlm.nih.gov/home/develop/api/} to perform query-based searches within PubMed database. The initial search retrieves a list of relevant PMIDs. Considering the long context length and inconsistent availability of full-text papers, our tool extracts only the titles and abstracts of these PMIDs. These elements are then concatenated and returned as the tool responses to provide retrieved context for fact-checking. In this paper, we limited the tool responses to a maximum of 5 papers per search and set the maximum number of iterative searches to 5.

\textbf{Agent Output.} The final output of \name{} is a structured \textbf{Report} consisting of four main sections: \textbf{Supporting Evidence}, \textbf{Refuting Evidence}, \textbf{Summary}, and \textbf{Conclusion}. The Supporting/Refuting Evidence section details the literature that either supports or refutes the claim, with each entry providing the PubMed ID (PMID), a verbatim Top Sentence quote from the abstract, and the relevant context. The Summary section outlines the overall reasons to support or refute the claim, accompanied by a consensus check and a final justification. Finally, the Conclusion section clearly presents the final verdict by outputting either `SUPPORTED' or `REFUTED' within `<answer>' tags. Details about the LangGraph implementation of \name{} with corresponding prompts are provided in Appendix~\ref{prompt}.

\subsection{Evidence-Grounded GRPO for \name{}}
In this section, we introduce Evidence-Grounded GRPO by first formalizing the task, detailing our task-specific reward design, and finally presenting the optimization objective.

\subsubsection{Task Definition}
We define our task of biomedical fact-checking reports generation with \name{} as a \textbf{Partially Observable Markov Decision Process (POMDP)} $(\mathcal{S}, \mathcal{A}, \mathcal{O}, \mathcal{T}, \mathcal{R})$ \cite{xi2024agentgymevolvinglargelanguage, xi2025agentgym}, where $\mathcal{S}, \mathcal{A}, \mathcal{O}, \mathcal{T}, \mathcal{R}$ represents the state space, the action space, the observation space, the state transition function and the reward function, respectively. Given a biomedical claim $c$, the LLM-based \name{} initializes at state $s_1$, which comprises the claim paired with an instructional prompt for fact-checking. At each time step $t$, given the current state $s_t \in \mathcal{S}$, the agent generates an action $a_t \sim \pi_\theta(\cdot|s_t)$ based on the policy model $\pi_\theta$ to process the task. The action $a_t \in \mathcal{A}$ here includes both the reasoning process and the search tool calling with a generated query. Upon executing the action, the agent receives the tool response as the observation $o_t \in \mathcal{O}$ from the environment, and the system transitions to the next state $s_{t+1} = \mathcal{T}(s_t, a_t)$. After $N$ interaction turns, the agent concludes the task by generating a fact-checking report as its final action $a_N$. A reward function $\mathcal{R}$ then evaluates this report to assign a final reward $R = \mathcal{R}(a_N)$.

\subsubsection{Task-Specific Reward Modeling}
EG-GRPO performs a task-specific reward design. The baseline approach of GRPO relies solely on a binary outcome-based reward, where $R=1$ if the agent task is performed successfully and $R=0$ otherwise. For \name{}, although an outcome-based reward, which evaluates whether the report's conclusion aligns with the ground truth label, can improve label prediction performance, generating high-quality biomedical fact-checking reports still requires more fine-grained feedback. Below, we detail the specific design for BioCheck reward $R_{\text{BioCheck}}$:

\textbf{Outcome-based Reward.} We retain the outcome-based reward, $R_{\text{outcome}}$, as a component of the final reward. Specifically, $R_{\text{outcome}} = 1.0$ if the final predicted label matches the ground truth provided by the dataset, and $R_{\text{outcome}} = 0.0$ otherwise.

\textbf{Advanced Search Reward.} To encourage \name{} to decompose the claim and perform advanced searches by connecting keywords with the Boolean operators `AND' or `OR', we introduce an additional reward, $R_{\text{search}}$, which is defined as the ratio of search queries containing Boolean operators to the total number of searches: $R_{\text{search}}=\frac{\text{Num of Searches with Boolean Operator}}{\text{Num of Searches}}$.

\textbf{Composite Evidence Reward.}
One critical factor of a fact-checking report is the quality of the evidence it contains. For each piece of retrieved evidence, we apply an additional reward model to assign a confidence score $s$ evaluating the probability that the evidence supports or refutes the claim. Specifically, we frame this claim verification task as a multiple-choice question, utilizing a lightweight LLM Qwen3.5-4B \cite{qwenmodel} to compute the token probability of selecting the supported or refuted answer based on the provided evidence. The prompt for the multiple-choice question and the corresponding implementation details to compute the evidence confidence score are included in Appendix~\ref{prob}.

Normally, multiple pieces of evidence are included for each report. To ensure both the quantity and quality of the retrieved evidence, we propose the Composite Evidence Reward. Formally, given a list of claim evidence pairs $(\mathcal{C},\mathcal{E}) = [(c_1,e_1),..., (c_N,e_N)]$, a label $y \in \{\text{supported},\text{refuted}\}$, and a reward model $\mathcal{L}$, we compute the probability score as $s_k = L(y|c_k,e_k)$. Then we compute the Composite Evidence Reward as
\begin{equation}
    R_{\text{CER}}((\mathcal{C},\mathcal{E}), y) = \frac{1}{N}\sum^{N}_{i=1}s_i(1-e^{-\alpha N})
\end{equation}
where the $\alpha$ coefficient serves as a scaling factor that controls the saturation rate: a larger value would cause the function to saturate more rapidly, meaning fewer pieces of evidence would be required to reach full reward. Here we set $\alpha=1.0$ by default.

The Composite Evidence Reward is computed for all abstracts retrieved by PMIDs. To ensure that the generated sentences serve as meaningful evidence, we only count the probability score when the corresponding abstract's probability score exceeds 0.5; otherwise, we set $s=0.0$. Notably, within each fact-checking report, the agent is allowed to retain evidence that contradicts its final decision, and the corresponding evidence scores are also included. This design incentivizes the model to critically retrieve evidence from both sides of a claim. Given the abstract evidence list $\mathcal{E}_{\text{supported}}, \mathcal{E}_{\text{refuted}}$, the evidence reward is defined as
\begin{equation}
    R_{\text{evidence}} = R_{\text{CER}}((\mathcal{C},\mathcal{E}_{\text{supported}}), \text{supported}) + R_{\text{CER}}((\mathcal{C},\mathcal{E}_{\text{refuted}}), \text{refuted})
\end{equation}

\textbf{Hallucination Penalty.} The report generation process is susceptible to hallucinations, such as the fabrication of PMIDs or Top Sentences in the generated evidence list. To mitigate this, we implement a rule-based hallucination detection mechanism that penalizes hallucinated evidence by assigning it a penalty score. We detect the evidence as a hallucination under two specific conditions: (1) if a provided PMID fails the Exact Match criterion within the retrieved PMID list; (2) if the generated Top Sentence yields a ROUGE-L precision score of less than 0.85 against the source abstract. Then we compute the accumulated hallucination penalty as the ratio of detected hallucinations to the total number of evidence:  $R_{\text{hallucination}}=\frac{\text{Num of Hallucination}}{\text{Num of Total Evidence}}$.


Overall, the final BioCheck reward is presented in the following formulation:
\begin{equation}
R_{\text{BioCheck}} = R_{\text{outcome}} + \beta_1 R_{\text{search}} + \beta_2 R_{\text{evidence}}-\beta_3R_{\text{hallucination}}
\end{equation}
Here, $\beta_1$, $\beta_2$ and $\beta_3$ are hyperparameters that control the weight of the extra rewards. We set $\beta_1=0.5$ and $\beta_2=\beta_3=1.0$ by default.

\subsubsection{EG-GRPO Formulation}
EG-GRPO shares the same objective function as the baseline GRPO \cite{shao2024deepseekmath}. Specifically, for each claim $c$ from the training claim set $C$, we first sample a group of trajectories $\{\tau_1, ..., \tau_G\}$ from the old policy $\pi_{\theta_{old}}$, with each trajectory $\tau_i = \{s_{i,1},a_{i,1},...,s_{i,|\tau_i|},a_{i,|\tau_i|}\}$ includes $|\tau_i|$ states and generated actions. Then, the policy model is optimized by maximizing the following objective function:
\begin{align}
\mathcal{J}_{\text{EG-GRPO}}(\theta) & = \mathbb{E}{[c \sim C, \{\tau_i\}_{i=1}^G \sim \pi_{\theta_{old}}(\tau|c)]} \\ 
& \frac{1}{G}\sum_{i=1}^G\frac{1}{|\tau_i|}\sum_{j=1}^{|\tau_i|} \big(\frac{1}{|a_{i,j}|}\sum_{t=1}^{|a_{i,j}|}
\{\min[w_{i,j,t}\hat{A}_{i,t},  \text{clip}\left(w_{i,j,t},1-\epsilon, 1+\epsilon\right)\hat{A}_{i,t}] -\beta\mathbb{D}_{\text{KL}} \}\big), \nonumber
\end{align}
where $w_{i,j,t} = \frac{\pi_\theta(a_{i,j,t}|s_{i,j},a_{i,j,<t})}{\pi_{\theta_{old}}(a_{i,j,t}|s_{i,j},a_{i,j,<t})}$ is the importance sampling term, $\epsilon$ is the clip ratio and $\beta$ is the ratio of KL penalty between policy model $\pi_\theta$ and reference model $\pi_{ref}$. $\hat{A}_{i,t}$ is the advantage calculated based on the relative rewards of the output reports inside each group. Given rewards $\mathbf{r} = \{R_1, R_2, ..., R_G\}$ of the output reports under the same group, the advantage is defined as $\hat{A}_{i,t}=\frac{R_i-\text{mean}(\mathbf{r})}{\text{std}(\mathbf{r})}$.

%% file: 4_experiment.tex
\section{Experiment}
In this section, we detail the experiments conducted to evaluate \name{} and EG-GRPO. Specifically, we outline the experimental settings, introduce the baseline methods, and discuss the final results with additional ablation study.

\subsection{Experimental Settings}
\textbf{Datasets.} 
We utilize two datasets for our experiments: SciFact \cite{wadden2020fact} and HealthFC \cite{vladika2024healthfc}. Because only SciFact provides explicit training, validation, and test splits, we used it for both training and evaluation, while using HealthFC only for evaluation. Furthermore, although both datasets include a `Not Enough Information' (NEI) label, its meaning differs between the two datasets. In SciFact, NEI indicates an absence of evidence to support or refute a claim specifically within the provided corpus. Due to the limited size of this corpus, it does not necessarily imply a lack of information in real world. Conversely, in HealthFC, where the data is collected from human written fact-checking articles, NEI means that there is no existing evidence about the claim in real word, or that the available evidence is inherently conflicting. For simplicity, our default experiments perform training and evaluation only on examples with binary labels `supported' and `refuted'. Finally, because the claims in HealthFC are originally formatted as questions, we utilized an advanced large language model (GPT-5.2) to convert them into declarative statements, ensuring structural alignment with SciFact.

\textbf{Base Models.} 
We evaluate \name{} across different LLM backbones, including the proprietary GPT-5.2 \cite{gpt5_2} and the open-source Qwen3.5-4B \cite{qwenmodel}. For the EG-GRPO, we specifically perform the training on Qwen3.5-4B with SciFact training split.

\textbf{Training Details.}
We implement EG-GRPO with the verl RL training library \cite{sheng2025hybridflow}. To adapt our agent for training, we integrate verl with the LangGraph framework to perform the rollout process. We apply EG-GRPO on the SciFact training set for 10 epochs with a learning rate of 1e-6. At least 4xNVIDIA A100 80GB GPUs are required to perform the training. During training, we perform evaluation on the validation set every 5 steps and select the checkpoint with the highest average reward as our final model. Additional hyperparameter details and training curves are provided in Appendix~\ref{training}.

\textbf{Task Evaluation.}
We evaluate \name{} on SciFact test set and HealthFC with various metrics for both label prediction and report generation tasks. (1) \textbf{Label Prediction:} Following previous fact-checking works \cite{barone2025combining, xie2025fire}, we evaluate label prediction by computing the accuracy (Acc), macro precision (Prec), macro recall (Rec), and macro F1 score (F1). For \name{}, we extract the predicted labels `SUPPORTED', `REFUTED', or `NOT ENOUGH INFO' from the Conclusion section of the generated report. Then, these labels are evaluated against the ground-truth labels provided in the dataset. (2) \textbf{Report Generation:} To evaluate report generation, we propose two novel metrics: Evidence Quality Score (EQS) and Evidence Hallucination Rate (EHR). The EQS is calculated by averaging the mean score of the evidence within each report that aligns with the report's conclusion, across all examples. The EHR is defined as the ratio of hallucinated instances to the total number of generated reports. Higher EQS and lower EHR values indicate greater report quality.

\textbf{Baseline Methods.}
We evaluate the performance of \name{} compared with the following baselines: (1) \textbf{No Retrieval}: A naive baseline directly applies an LLM for label prediction without any external information. (2) \textbf{CER} \cite{barone2025combining}: LLM classification with context provided by embedding-based retrieval. (3) \textbf{FIRE} \cite{xie2025fire}: An LLM-based agent equipped with Google Search. (4) \textbf{PMSearch Agent}: An LLM-based LangGraph agent shares the same architecture as \name{}, but outputs only the predicted label after conducting a search, rather than a comprehensive fact-checking report. (5) \textbf{\name{} with GRPO}: For comparison with EG-GRPO, we also implement a baseline GRPO on \name{} without the task-specific reward design.

While all baselines are evaluated on the label prediction task, existing literature lacks established benchmarks for evaluating fact-checking report generation. Therefore, our evaluation on report generation focuses primarily on comparing the base model against those optimized with baseline GRPO and EG-GRPO. Further implementation details for all baselines are provided in Appendix~\ref{baseline}.

\subsection{Main Results}
\textbf{Label Prediction.} We first present the Label Prediction results for both the SciFact and HealthFC in Table~\ref{tbl1}. As shown in the table, \name{} optimized with EG-GRPO consistently outperforms all baselines under the Qwen3.5-4B base model, including training-free agents and \name{} trained with baseline GRPO. Notably, applying EG-GRPO to the lightweight, open-source Qwen3.5-4B model enables it to surpass even the proprietary GPT-5.2 model, achieving an 8.9\% improvement in prediction accuracy on SciFact. Furthermore, although not explicitly trained on the HealthFC dataset, our model still exhibits improved performance. While the baseline GRPO also improves \name{} performance on SciFact, it yields limited accuracy gains and a worse F1 score on HealthFC. This demonstrates that relying solely on an outcome-based reward causes the model behavior to easily overfit to the training dataset. In contrast, our EG-GRPO, with its evidence-aware reward design, achieves better transferability in label prediction by optimizing the report quality.

\begin{table}[ht]
\centering
\caption{Label prediction performance of \name{} against various baselines. For each Base Model, the highest value is bolded and the second highest is underlined.}
\label{tbl1}
\resizebox{0.95\textwidth}{!}{
\begin{tabular}{ll cccc cccc}
\toprule
\multirow{2}{*}{\textbf{Base Model}} & \multirow{2}{*}{\textbf{Method}} & \multicolumn{4}{c}{\textbf{SciFact}} & \multicolumn{4}{c}{\textbf{HealthFC}} \\
\cmidrule(lr){3-6} \cmidrule(lr){7-10} 
 & & Acc & Prec & Rec & F1 & Acc & Prec & Rec & F1 \\
\midrule
\multirow{5}{*}{GPT5.2} 
 & No Retrieval & 66.49 & 84.41 & 65.47 & 71.74 & 60.86 &\textbf{84.22} & 53.68&59.52 \\
 & CER \cite{barone2025combining}          & 78.53 & \textbf{96.25} & 78.73 & \textbf{86.47} & 54.13 & \underline{82.95} & 47.01 & 53.00 \\
 & FIRE \cite{xie2025fire} & \textbf{82.72} & 85.59 & \textbf{83.54} & 82.57 & \textbf{74.01} & 72.58 & \textbf{73.01} & \underline{72.76} \\
 & PMSearch Agent& 78.01 & \underline{92.12} & 78.24& 84.43 & 70.95&77.36 &67.79& 71.78\\
 & \textbf{\name{}} &\underline{81.15} &88.98 & \underline{81.57} & \underline{84.68} & \underline{72.48} & 75.34 & \underline{71.47} & \textbf{73.34} \\
\midrule
\multirow{7}{*}{Qwen3.5-4B} 
 & No Retrieval &59.16 & 72.11 &58.66 &64.45 &50.15 & 64.56&44.56 &50.59 \\
 & CER \cite{barone2025combining}         & 75.92 & 88.86 & 75.84 & 81.80 & 52.60 & 73.88 & 47.76 & 56.23 \\
 & FIRE \cite{xie2025fire} &81.15 & 82.33 & 81.69 & 81.11 & 71.87 & 70.50 & \underline{70.37} & \underline{70.43} \\
 & PMSearch Agent&81.15 &90.01 &81.09 &85.32 &70.64& \underline{76.29} &66.48 &69.90 \\
\cmidrule(lr){2-10}
 & \textbf{\name{}} &80.10 &89.22 &80.40 &84.30 & 70.34& 74.91&66.99 &70.28 \\ 
    & w/ GRPO &\underline{89.01} & \underline{90.27} &\underline{88.76} &\underline{89.38} & \underline{72.17} & 74.45 & 65.73 & 66.10\\
 & w/ \textbf{EG-GRPO} &\textbf{90.05} & \textbf{91.02} &\textbf{90.17} &\textbf{90.51} & \textbf{77.37} & \textbf{79.51} & \textbf{72.38} & \textbf{73.59}\\
\bottomrule
\end{tabular}
}
\end{table}

When comparing the results among training-free agents, we found that \name{} shows only marginal improvements in accuracy and F1 score compared to the PMSearch Agent on GPT-5.2, and exhibits even worse performance under Qwen3.5-4B. This is likely because report generation introduces additional complexity and requires advanced model capabilities to execute effectively. We also noted that the baseline method CER, which applies embedding-based retrieval within the PubMed database and re-ranking before justification, achieves high performance on SciFact but performs poorly on HealthFC. This discrepancy likely occurs because SciFact claims are sourced directly from PubMed, making embedding-based retrieval highly effective in-domain, whereas it struggles to process the more general, out-of-domain health claims in HealthFC. Furthermore, when comparing the two base models, we observe the most substantial performance disparity in the No Retrieval setting. This demonstrates that while GPT-5.2 possesses superior internal parametric knowledge, the introduction of external evidence retrieval effectively bridges this gap.

\begin{figure}[htbp]
  \centering
  \begin{subfigure}{0.45\columnwidth}
    \centering
    \includegraphics[width=\linewidth]{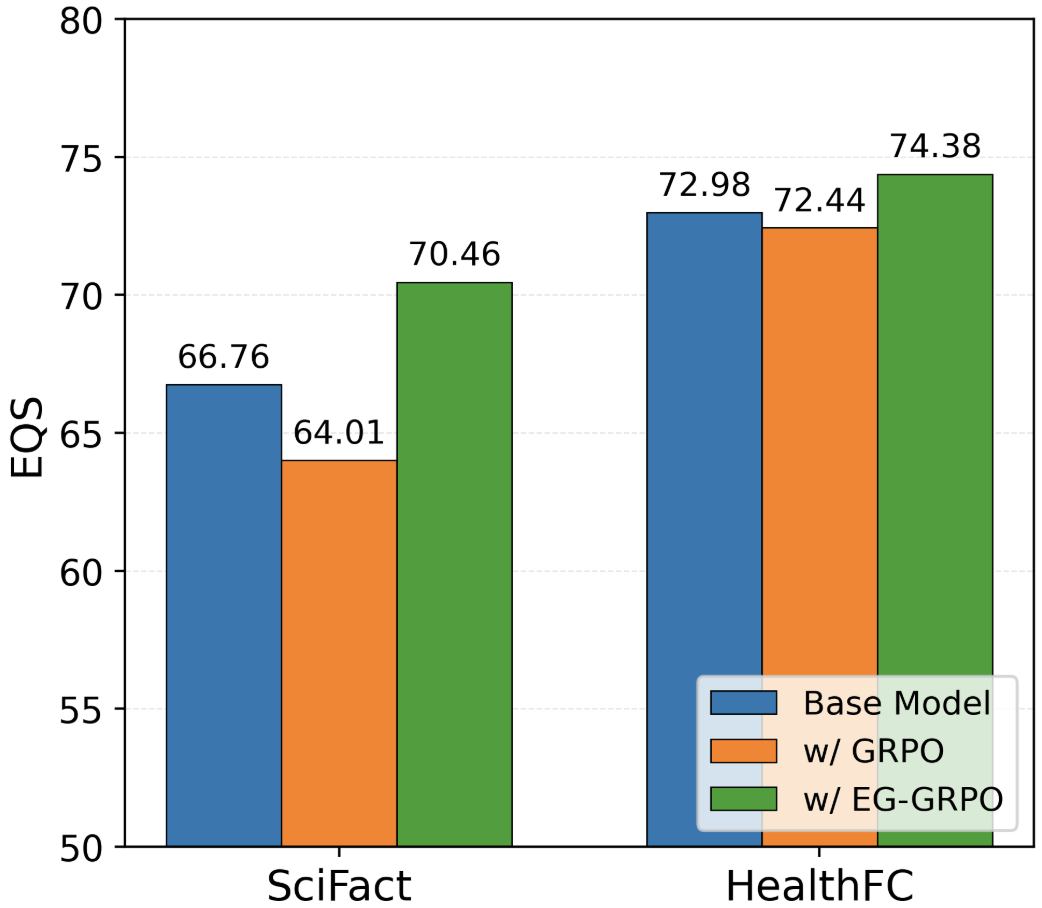}
    \caption{}
    \label{fig:sub3}
  \end{subfigure}\hfill
  \begin{subfigure}{0.45\columnwidth}
    \centering
    \includegraphics[width=\linewidth]{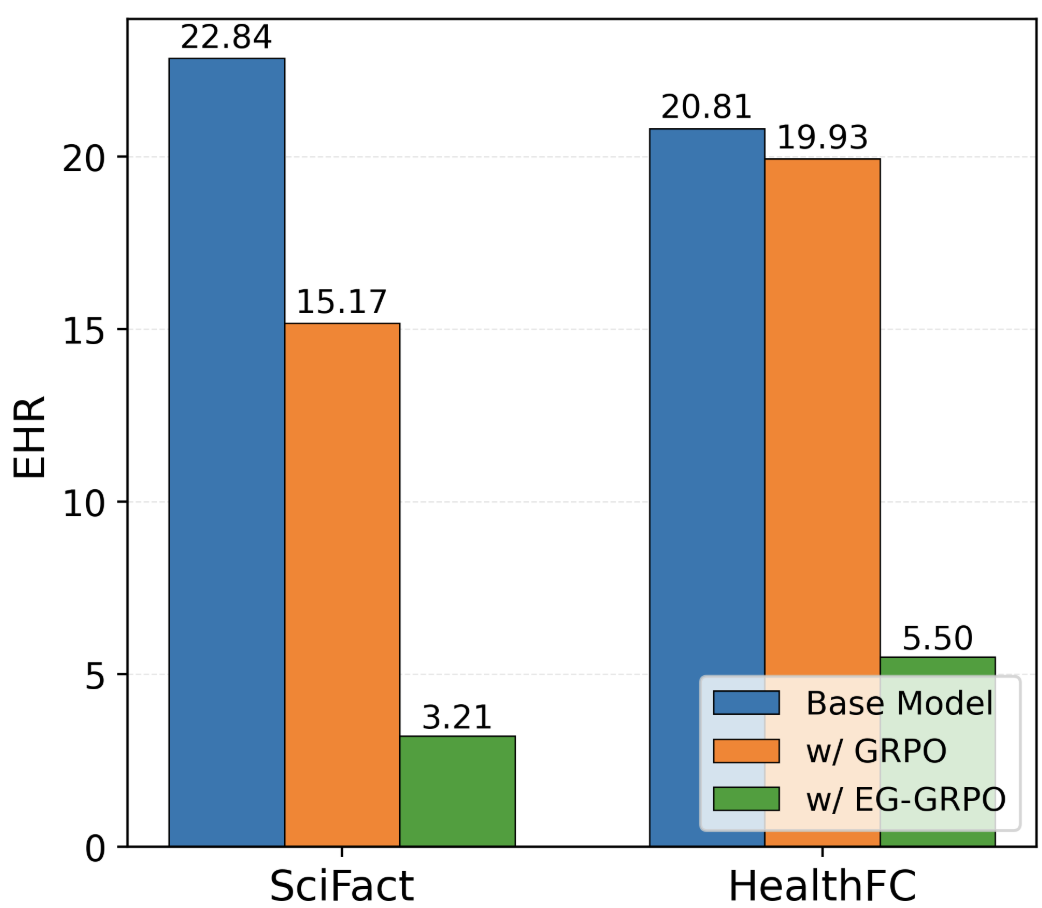}
    \caption{}
    \label{fig:sub4}
  \end{subfigure}
  \caption{Sub-figures (a) and (b) evaluate the EQS (\%) and EHR (\%), respectively, for the generated fact-checking reports across various models.}
  \label{fig:narrow-side-by-side}
\end{figure}

\textbf{Report Generation.} For the report generation task, we evaluate the Evidence Quality Score (EQS) and Evidence Hallucination Rate (EHR) respectively in Figure~\ref{fig:sub3} and Figure~\ref{fig:sub4} on both the SciFact and HealthFC datasets with the base model Qwen3.5-4B. As shown in the figures, compared to \name{} with the base model and baseline GRPO, \name{} optimized with EG-GRPO generates biomedical fact-checking reports with a higher EQS and a significantly reduced EHR. This reflects higher average evidence quality with fewer hallucinations, demonstrating the effectiveness of EG-GRPO in improving the overall quality of biomedical fact-checking reports. Additionally, we find that \name{} with baseline GRPO results in both lower EQS and EHR. This demonstrates that without the specific reward design used in EG-GRPO, baseline GRPO still impacts report generation quality when optimizing the outcome-based reward. However, lacking specific guidance, it yields only a marginal reduction in EHR while suffering from an even lower EQS. Combined with the results in Table~\ref{tbl1}, we demonstrate that our EG-GRPO can improve \name{} in both label prediction and report generation performance.

\textbf{Qualitative Analysis of Generated Reports.} To provide a better understanding of the quality of generated reports, Appendix~\ref{example} includes two sample reports: one generated by the base model (Qwen3.5-4B) and another by the model optimized with EG-GRPO. A comparison of these two examples reveals a clear difference in quality. The base model cited two mismatched PMIDs and retrieved irrelevant evidence, ultimately leading to an erroneous conclusion. In contrast, \name{} optimized by EG-GRPO retrieved evidence with much higher quality. After a critical analysis of the context, it finally made the correct judgment.


\subsection{Ablation Study}
\textbf{Evidence Confidence Score with Alternative Judge LLM.} In our default setting, we use the same Qwen3.5-4B model as the judge LLM to compute the evidence confidence score during training with EG-GRPO and for the EQS evaluation. To demonstrate that the improvements in EQS do not rely on the specific judge LLM used during training, we perform additional evaluations using an alternative LLM, GPT-4o \cite{hurst2024gpt} to compute the evidence confidence score. GPT-4o is currently the most advanced GPT model that provides logprob outputs. The prompt and implementation are identical to those used for the Qwen3.5-4B model, as detailed in Appendix~\ref{prob}. We present the EQS results using GPT-4o as the judge in Table~\ref{tbl2}. The highest EQS among the various agent models is highlighted in bold.

\begin{table}[ht]
\centering
\caption{EQS (\%) evaluation with alternative judge LLM.}
\label{tbl2}
\resizebox{0.6\textwidth}{!}{
\begin{tabular}{l cc cc}
\toprule
\multirow{2}{*}{\textbf{Agent Model}}  & \multicolumn{2}{c}{\textbf{Qwen3.5-4B}} & \multicolumn{2}{c}{\textbf{GPT-4o}} \\
\cmidrule(lr){2-3} \cmidrule(lr){4-5} 
 & SciFact & HealthFC & SciFact & HealthFC \\
\midrule
 Qwen3.5-4B & 66.76 & 72.98 & 64.73 & 73.83 \\
 w/ GRPO   & 64.01 & 72.44 & 63.60 & 71.65   \\
 w/ EG-GRPO& \textbf{70.46} & \textbf{74.38} & \textbf{69.92} & \textbf{75.84} \\
\bottomrule
\end{tabular}
}
\end{table}

As shown in the table, although evaluating with GPT-4o yields different EQS values, the overall trend remains consistent: \name{} optimized with EG-GRPO continues to outperform the base model and the baseline GRPO in generating fact-checking reports with higher evidence quality.

%% file: 5_conclusion.tex
\section{Conclusion}
In this paper, we mitigate the pervasive issue of health misinformation by advancing automated biomedical fact-checking from a traditional text classification task to comprehensive report generation. Rather than merely outputting an isolated prediction label, fact-checking reports provide contextualized information, serving as a much better reference for human decision-makers. To achieve this, we propose \name{} to generate structured biomedical fact-checking reports by performing agentic search within the PubMed database. To further enhance the agent's performance, we propose an end-to-end RL method named Evidence-Grounded GRPO (EG-GRPO), which utilizes a task-specific reward to optimize the groundedness and quality of the generated evidence in reports. Comprehensive experiments demonstrate the efficacy of \name{} optimized with EG-GRPO, achieving both accurate label prediction and high-quality report generation for biomedical fact-checking, making it possible to assist humans in discerning health misinformation.

\section{Limitation}
While \name{} provides an effective automated approach for biomedical fact-checking, its current scope is primarily limited to isolated atomic claims. In real-world scenarios, health misinformation is rarely presented in such a clean format. Instead, it is normally embedded within complex narratives across diverse platforms (e.g., social media posts, news articles) and often relies on multi-modal contexts, such as images and videos. Extending \name{} to verify claims within these environments remains a significant challenge for future work.

A second limitation concerns the lack of high-quality data for training our agent. In fact, the SciFact training set contains only around 500 labeled examples and lacks Not Enough Info (NEI) cases that frequently occur in the wild. Furthermore, because SciFact is constructed from scientific claims extracted from PubMed abstracts, there is a notable distribution shift when applying the agent model to general public health. This gap largely explains why \name{} presents a comparably modest performance improvement on the HealthFC dataset compared to SciFact.

%% file: appendix.tex
\section{Implementation Details of \name{}} \label{prompt}
This section presents how we use the LangGraph framework to build \name{}.

\begin{wrapfigure}{r}{0.5\textwidth}
    \centering
    \vspace{-26pt}
    \includegraphics[width=0.15\textwidth]{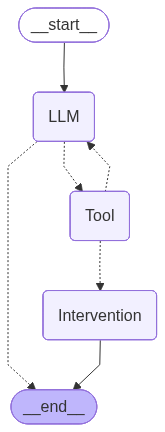}
    \caption{LangGraph structure of \name{}.}
    \label{fig:langgraph}
\end{wrapfigure}

\subsection{LangGraph}
A demonstration figure for the structure of our LangGraph agent is presented in Figure~\ref{fig:langgraph}.

Based on the LangGraph structure shown in the figure, \name{} starts at an `LLM' node that processes the initial prompts. A conditional edge then directs the flow: if tool calls are detected in LLM outputs, it triggers the `Tool' node; if not, the process ends. The `Tool' node then executes the generated tool calls in LLM outputs and tracks the number of tool calls. If the number exceeds a maximum tool call limit, the flow is directed to an `Intervention' node where an extra user prompt is injected to force a conclusion. If under the limit, the agent cycles back to the LLM for further iteration.

\subsection{Prompt}
We outline the prompts utilized in \name{}, including a system prompt that defines the agent's role and task; an initial user prompt containing the core task instructions; and an intervention prompt (injected as a user message), which triggers when the maximum search limit is reached to terminate searching and force the generation of the final fact-checking report.

\textbf{\textit{System Prompt}}
\begin{mybox}
You are a biomedical expert. Your goal is to analyze a provided claim with the PubMed database by using the `search\_pubmed' tool. Based on your search results, synthesize your findings into a brief report and determine the most appropriate label: supported or refuted.
\end{mybox}

\textbf{\textit{Initial User Prompt}} to start the agent. The red-highlighted text indicates a placeholder for the claim undergoing fact-checking.
\begin{mybox}
Follow this strict protocol to verify the claim:\\

**AGENT PROTOCOL**\\
1. Analyze \& Retrieve: Analyze the claim to identify Core Entities (with expanded aliases/MeSH terms), the Relation, and the Context. Construct a comprehensive Boolean query using advanced operators (AND/OR) connecting these keywords, and use the `search\_pubmed' tool to execute your initial search.\\
2. Iterate with Adaptive Strategies: If the search yields insufficient evidence, iteratively refine your query and continue searching until you collect sufficient evidence or hit the maximum number of searches (5). Apply the following strategies dynamically during your refinement process:

\ \ \ \  - Constraint Relaxation: Drop Context or Relation limits if you get 0 hits or too little evidence.

\ \ \ \  - Adaptive Incorporation: Extract newly discovered synonyms, related pathways, or methodologies from abstracts to formulate more precise follow-up queries.

\ \ \ \  - Active Refutation: Actively search for contradictory evidence (e.g., append "fails to", "opposite", or "artifact") if you find 0 supporting hits or are stuck in a one-sided confirmation loop.\\
3. Final Report \& Conclusion: Once you have gathered sufficient evidence to make a definitive conclusion, you MUST output your findings STRICTLY in the format below:\\

\#\#\# REPORT\\
**Supporting Evidence:**\\\
[For each supporting article, list:]\\
- PMID: [e.g., 12345678]\\
- Top Sentence: [Direct, verbatim quote from the abstract enclosed in double quotation marks to support the claim]\\
- Context: [Note any specific conditions, e.g., "Only in murine models"]\\
*(If none found, write: "No supporting evidence found.")*\\

**Refuting Evidence:**\\\
[For each refuting article, list:]\\
- PMID: [e.g., 87654321]\\
- Top Sentence: [Direct, verbatim quote from the abstract enclosed in double quotation marks to support the claim]\\
- Context: [Note any specific conditions]\\
*(If none found, write: "No refuting evidence found.")*\\

**Summary**\\
- Reasons to Support: [2-3 sentences analyzing and summarizing findings to support the claim]\\
- Reasons to Refute: [2-3 sentences analyzing and summarizing findings to refute the claim]\\
- Consensus \& Context Check: [Evaluate if the claim's scope matches the evidence (e.g., is it a universal truth or a special case?)]\\
- Final Justification: [1-2 sentences explaining the final decision]\\

**Conclusion**\\\
[Output STRICTLY ONE of the tags below on a new line]\\
<answer>SUPPORTED</answer>\\
OR\\
<answer>REFUTED</answer>\\ \\

Please verify this claim now: \textcolor{red}{CLAIM}
\end{mybox}

\textbf{\textit{Intervention Prompt}}
\begin{mybox}
Attention: You have reached the maximum step limit. Do not call any more tools. Please provide the final report with conclusion based on the information collected so far. If the current information is insufficient to make the conclusion, please respond with `<answer>NOT ENOUGH INFORMATION</answer>' as your final conclusion after the report.
\end{mybox}

\section{Computing the Evidence Confidence Score} \label{prob}

To compute the Composite Evidence Reward and evaluate the Evidence Quality Score, we use an LLM to calculate the evidence confidence score that the provided evidence supports or refutes the claim. Since this is a classification task rather than text generation, we frame it as a multiple-choice problem and compute the probability of each choice's token. Specifically, we prompt the LLM to output a choice from `A: supported', `B: refuted', or `C: not enough information' based on the provided evidence and claim. Instead of answer generation, we only consider the next token's logprob outputs $z$ and compute the probability score $s = \exp(z)$ for the corresponding token `A', `B', or `C'.

The prompt for the LLM is shown below. Here, the red-highlighted `EVIDENCE' and `CLAIM' serve as placeholders.

\begin{mybox}
Given the abstract and the claim, decide whether the claim is supported, refuted by the abstract, or if there is not enough information to make a conclusion.\\

Abstract: \textcolor{red}{EVIDENCE}\\

Claim: \textcolor{red}{CLAIM}\\

Return exactly one choice for the claim verification result: `A: supported', `B: refuted', or `C: not enough information'.
\end{mybox}

\section{Training Details} \label{training}
This section includes the training details of applying EG-GRPO on \name{}. 

Regarding the reward modeling hyperparameters, we utilize the default values defined in Section 3.2.2. Specifically, we set the saturation factor $\alpha=1.0$ for $R_{\text{CER}}$. For the Biocheck Reward, formulated as $R_{\text{outcome}} + \beta_1 R_{\text{search}} + \beta_2 R_{\text{evidence}} - \beta_3 R_{\text{hallucination}}$, we set $\beta_1=0.5$ and $\beta_2=\beta_3=1.0$.

The key hyperparameters configured for the verl reinforcement learning library are detailed in the following Table~\ref{tab:hyperparameters}.

\begin{table}[htbp]
    \centering
    \caption{Hyperparameters for EG-GRPO.}
    \label{tab:hyperparameters}
    \begin{tabular}{c|c}
        \toprule
        \textbf{Hyperparameter} & \textbf{Value} \\
        \midrule
        total epochs & 10 \\
        global batch size & 64 \\
        mini batch size & 64 \\
        learning rate & 1e-6 \\
        KL loss coefficient & 0.03 \\
        max response length & 16000 \\
        clip ratio low & 0.2 \\
        clip ratio high & 0.28 \\
        num of agents in group & 8 \\
        \bottomrule
    \end{tabular}
\end{table}

\begin{figure}[ht]
    \centering
    \includegraphics[width=0.95\textwidth]{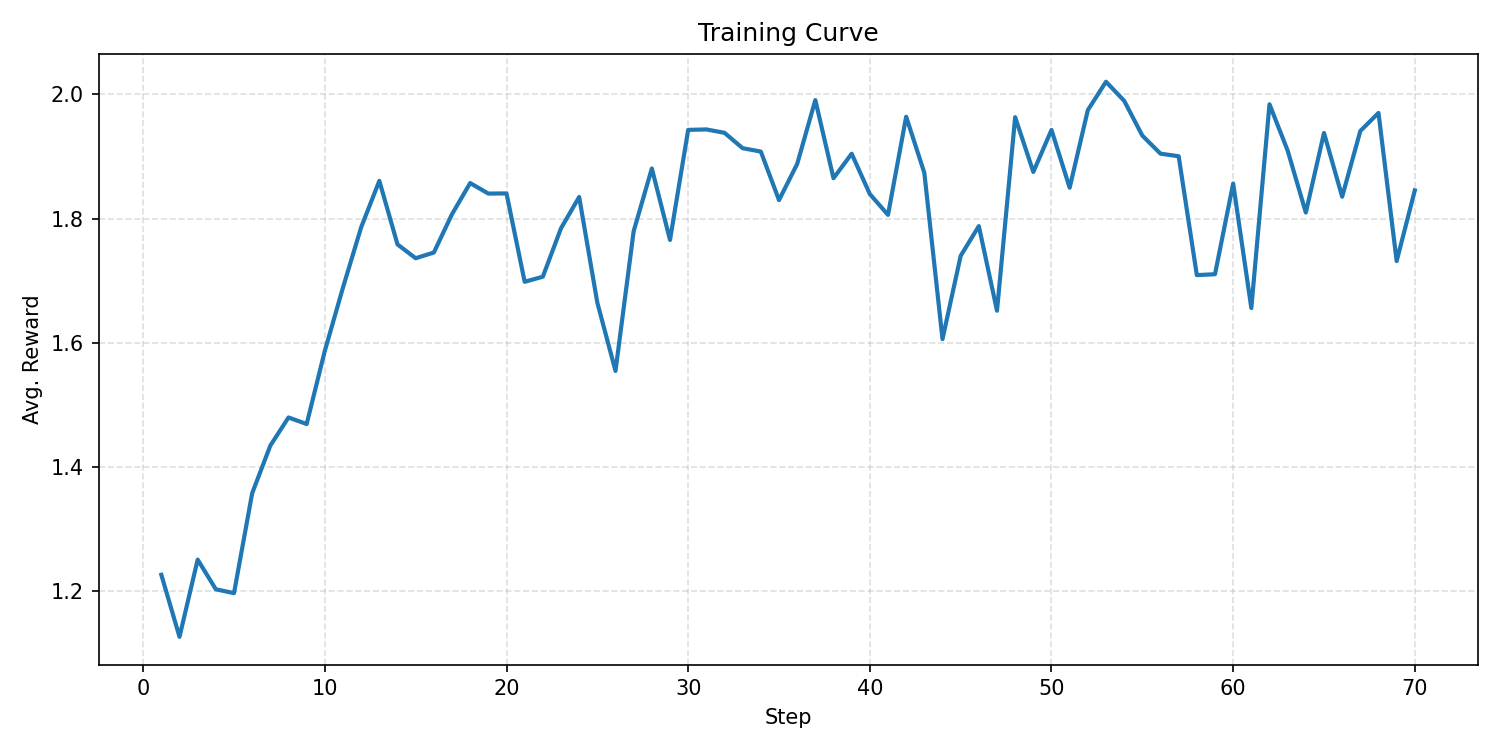}
    \caption{Training Curve for EG-GRPO.}
    \label{fig:training}
\end{figure}

The entire training process has 70 steps, with the corresponding training curve depicted in Figure~\ref{fig:training}. We select the checkpoint at step 60 as our final agent model, as it achieves the highest average reward on the validation set.

\section{Baseline Methods} \label{baseline}
In this section, we detail the implementation of the baseline methods shown in Table~\ref{tbl1}.

\textbf{No Retrieval.} 
As a naive baseline, we prompt the LLM to fact-check the claim using only its internal knowledge. The corresponding system and user prompts are provided below, with the placeholder for the claim highlighted in red.

\textbf{\textit{System Prompt}}
\begin{mybox}
You are a biomedical expert. Your goal is to analyze a provided claim based on your biomedical expertise and classify it using one of three labels.
\end{mybox}

\textbf{\textit{User Prompt}}
\begin{mybox}
To verify the claim, directly provide your classification result exactly as `<answer>SUPPORTED</answer>', `<answer>REFUTED</answer>', or `<answer>NOT ENOUGH INFORMATION</answer>'. \\ \\

Please verify this claim now: \textcolor{red}{CLAIM}
\end{mybox}


\textbf{CER \cite{barone2025combining}.} 
We implement Combining Evidence and Reasoning (CER) as a retrieval-augmented biomedical fact-checking baseline following an evidence--reasoning--classification pipeline. For each input claim, we retrieve candidate evidence from an external PubMed abstract corpus using a sparse BM25 retriever \cite{robertson2009probabilistic}. The retrieval index is built over PubMed abstracts with standard text preprocessing, including lower-casing, stemming, stop-word removal, and normalization of punctuation, special characters, and acronyms.

Given a claim, the retriever first returns the top 20 PubMed abstracts. Each retrieved abstract is then segmented into sentences, and candidate evidence sentences are reranked using a biomedical sentence encoder. We encode the claim and candidate sentences in the same embedding space and select the top 3 sentences from each retrieved abstract according to cosine similarity. This step yields an evidence set of up to 60 PubMed sentences per claim.

The original CER framework prompts an LLM with retrieved scientific evidence for evidence-grounded reasoning and then trains a supervised classifier to produce the final veracity prediction. Our implementation follows the same retrieval and reasoning design but replaces the trained classifier with an LLM for zero-shot decision. In the first stage, the LLM is prompted as a biomedical reasoning model and is provided only with the retrieved PubMed evidence. It generates a preliminary decision together with a concise justification grounded in the retrieved evidence. We follow the prompt design of the original CER framework for this reasoning stage and refer readers to \cite{barone2025combining} for the full prompt specification. In the second stage, the LLM receives the claim, retrieved evidence, preliminary decision, and generated justification to output the final decision. In summary, compared with the original CER framework, our implementation preserves the evidence retrieval and reasoning components but uses an LLM, rather than a trained classifier, to make the final veracity decision.

\textbf{FIRE \cite{xie2025fire}.}
We use Fact-checking with Iterative Retrieval and Verification (FIRE) as an iterative retrieval-and-verification baseline for atomic claims. Unlike fixed-depth retrieval pipelines, FIRE allows the verifier LLM to adaptively determine whether the currently available evidence is sufficient for a final factuality judgment or whether additional evidence should be retrieved. At each iteration, the model receives the claim together with the accumulated evidence snippets and either returns a final decision or issues a new Google search query. We follow the original FIRE prompting protocol for iterative verification and query generation and refer readers to \cite{xie2025fire} for the full system and user prompts.

When additional evidence is requested, we retrieve web evidence through Google Search using the Serper API and collect the top 3 results for each query. The evidence pool is constructed from available search-result snippets, including answer-box content, knowledge-graph descriptions, and organic-result snippets. Thus, FIRE differs from corpus-based baselines in that it does not rely on a fixed offline evidence collection; instead, it verifies claims against evidence retrieved from the live web.

The iterative retrieval-and-verification loop runs for at most 5 steps, with up to 10 retries allowed when the model output does not match the required format. To reduce redundant retrieval, the loop can terminate early if the model repeatedly generates semantically similar queries or retrieves highly similar evidence. Semantic similarity is computed using \texttt{all-MiniLM-L6-v2} sentence embeddings with cosine similarity, with a threshold of 0.9. If the search budget is exhausted before a final decision is produced, FIRE performs a final verification step using all accumulated evidence. The resulting prediction is mapped to a binary factuality label.

\textbf{PMSearch Agent.} 
To ensure a fair comparison within the same LangGraph framework, we developed a baseline agent named PMSearch Agent. This baseline shares the same LangGraph structure as \name{}. It also utilizes the `pubmed\_search' tool to retrieve 5 relevant papers per search, with a maximum limit of 5 search iterations. The system prompt, initial user prompt, and intervention prompt for the PMSearch Agent are presented below.

\textit{\textbf{System Prompt}}
\begin{mybox}
You are a biomedical expert. Your goal is to determine if a provided claim is supported or refuted with the PubMed database. To access the PubMed database, you can use the `search\_pubmed' tool to retrieve relevant papers.
\end{mybox}

\textit{\textbf{Initial User Prompt}} with claim placeholder highlighted.
\begin{mybox}
Follow this strict protocol to verify the claim:\\

**AGENT PROTOCOL**\\
1. Retrieve Evidence with Search Tool: Use the `search\_pubmed' tool with targeted search query to find papers that can verify the claim.\\
2. Iterate if Necessary: If the search yields no results (0 hits) or the search result is insufficient to make a conclusion, iteratively refine your query and continue searching until you reach the final conclusion or hit the maximum number of searches (5).\\
3. Final Decision: Once you have gathered sufficient evidence to make a definitive conclusion, provide a brief explanation followed strictly by `<answer>SUPPORTED</answer>' or `<answer>REFUTED</answer>' as your final conclusion.\\ \\

Please verify this claim now: \textcolor{red}{CLAIM}
\end{mybox}

\textit{\textbf{Intervention Prompt}}
\begin{mybox}
Attention: You have reached the maximum step limit. Do not call any more tools. Please provide the best possible final answer based on the information collected so far. If the current information is insufficient to make a decision, please respond with `<answer>NOT ENOUGH INFORMATION</answer>'.
\end{mybox}

\textbf{\name{} with GRPO.} We compare our proposed EG-GRPO against a standard GRPO baseline applied to \name{}. The baseline GRPO relies solely on an outcome-based reward, evaluating whether the generated report's final conclusion matches the ground truth label. To ensure a fair comparison, all training parameters remain identical between the two approaches within the verl RL library. We apply GRPO to the SciFact training split, evaluating the average validation reward every 5 steps to identify the optimal checkpoint. For \name{} with GRPO, the highest average reward was achieved at step 60, which we consequently selected as our final agent model.

\section{Examples of Generated Fact-Checking Report} \label{example}
This section includes two examples of biomedical fact-checking reports generated by \name{} with the base model Qwen3.5-4B and the model optimized with EG-GRPO respectively. The claim to verify is "Azo food dyes cause attention deficit disorder and hyperactivity in children." with the ground truth label "SUPPORTED". Red colored text within brackets indicates warnings about the mismatched PMIDs and low quality evidence.

\textbf{Fact-Checking Report Generated by \name{} with the Base Model}

\begin{promptblock}
\#\#\# REPORT \\
**Supporting Evidence:**\\
- PMID: 17825405 \\
- Top Sentence: "Artificial colours or a sodium benzoate preservative (or both) in the diet result in increased hyperactivity in 3-year-old and 8/9-year-old children in the general population."\\
- Context: Randomized, double-blinded, placebo-controlled crossover trial in UK children\\

- PMID: 19882794 \\
- Top Sentence: "A meta-analysis of 15 double-blind clinical trials that evaluated artificial food colouring in children already considered to be hyperactive showed an increase in their hyperactive behavior."\\
- Context: Meta-analysis of 15 double-blind RCTs across studies evaluating food colorings\\

- PMID: 33953077 \quad\quad\quad \textcolor{red}{[PMID 33953077 does not exist within the retrieved evidence list.]}\\
- Top Sentence: "Forty (72.7\%) demonstrated improved behaviour and 26 (47.3\%) remained improved following liberalization of the diet... Two significant reactors were identified whose behavioural pattern featured extreme irritability, restlessness and sleep disturbance."\\
- Context: Double-blind crossover study on children with suspected hyperactivity\\

- PMID: 7169286 \quad\quad\quad \textcolor{red}{[Low evidence confidence score supporting the claim: $s=0.09$.]} \\
- Top Sentence: "The hyperactive children exhibited a greater degree of physiological activity following both the placebo and the drink containing food additives than did the control children. However, the magnitude of physiological changes in the hyperactive children were greater in response to ingestion of the additives than to placebo."\\
- Context: Randomized double-blind crossover study measuring heart rate and EEG in hyperactive children\\

- PMID: 6400044 \quad\quad\quad \textcolor{red}{[Low evidence confidence score supporting the claim: $s=0.05$.]}\\
- Top Sentence: "Only a subset of hyperactive children seem to be affected by food colors."\\
- Context: Review discussing individual sensitivity\\

**Refuting Evidence:**\\
- PMID: 8747098\\
- Top Sentence: "Numerous double-blind studies of the Feingold hypothesis have led to the rejection of the idea that this elimination diet has any benefit beyond the normal placebo effect."\\
- Context: Review of dietary elimination trials\\

- PMID: 24217026\\
- Top Sentence: "Capsule A containing AFCs and Capsule B containing sodium benzoate had no significant adverse effect compared with placebo in both behavior scores."\\
- Context: Randomized, double-blind, placebo-controlled trial in 130 Hong Kong children aged 8-9 years\\

- PMID: 6 \quad\quad\quad \textcolor{red}{[PMID 6 does not exist within the retrieved evidence list.]}\\
- Top Sentence: "When the best probably blinded assessment was employed, effects remained significant for free fatty acid supplementation... but were substantially attenuated to nonsignificant levels for other treatments."\\
- Context: Meta-analysis of dietary and psychological ADHD treatments\\

**Summary**\\
- Reasons to Support: Multiple controlled studies (especially meta-analyses and RCTs) demonstrate associations between artificial food colorings and increased hyperactivity/ADHD symptoms in children. Physiological measurements show differential responses to additives versus placebo in hyperactive children.\\

- Reasons to Refute: Several high-quality double-blind placebo-controlled trials found no significant association between food colorings and behavior in general population children (Hong Kong trial, Feingold diet review requiring blinded assessment). Some effects appear limited to subgroups of "sensitive" children.\\

- Consensus \& Context Check: The evidence is conflicting and inconsistent across studies. The claim of a universal causal relationship is not supported - effects appear to be variable, potentially limited to certain age groups or sensitive individuals rather than a blanket effect.\\

- Final Justification: Given the contradictory evidence from multiple quality-controlled studies, the claim that azo food dyes universally cause ADHD/hyperactivity in children cannot be definitively supported. The relationship appears limited to certain populations, though some studies do show associations.\\

**Conclusion**\\
<answer>REFUTED</answer>
\end{promptblock}

\textbf{Fact-Checking Report Generated by \name{} with EG-GRPO}
\begin{promptblock}
\#\#\# REPORT \\
**Supporting Evidence:** \\
- PMID: 38423749\\
- Top Sentence: "A systematic review has established that artificial azo dye food colourants (AFCs) and sodium benzoate preservative cause disturbed behaviour in children."\\
- Context: Systematic review of observational and animal model studies\\

- PMID: 15613992\\
- Top Sentence: "Meta-analytic modeling determined the overall effect size of AFCs on hyperactivity to be 0.283 (95\% CI, 0.079 to 0.488), falling to 0.210 (95\% CI, 0.007 to 0.414) when the smallest and lowest quality trials were excluded."\\
- Context: Analysis of 15 double-blind placebo-controlled trials; acknowledges publication bias\\

- PMID: 7965420\\
- Top Sentence: "Behavioral changes in irritability, restlessness, and sleep disturbance are associated with the ingestion of tartrazine in some children. A dose response effect was obtained."\\
- Context: Double-blind placebo-controlled study; significant reactions observed at all six dose levels tested (1-50mg tartrazine)\\

- PMID: 2857900\\
- Top Sentence: "Artificial colorants and preservatives were the commonest provoking substances, but no child was sensitive to these alone."\\
- Context: Oligoantigenic diet trial with 76 overactive children (62 improved)\\

**Refuting Evidence:**\\
- PMID: 20013441\\
- Top Sentence: "For both groups, mean intakes of the food additives among consumers only were far below the doses used in the previous study on hyperactivity. No child or teenager achieved the overall intakes used in the study linking food additives with hyperactivity."\\
- Context: Irish children and teenagers; dietary intake below threshold doses used in challenge studies\\

- PMID: 8277950\\
- Top Sentence: "Even when intake exceeds typical dietary levels, neither dietary sucrose nor aspartame affects children's behavior or cognitive function."\\
- Context: Double-blind trial in normal preschool and school-age children; no artificial food coloring in diets tested\\

**Summary**\\
- Reasons to Support: Multiple double-blind controlled studies and a meta-analysis demonstrate that a subgroup of children (approximately 23-89\% depending on regimen) experience irritability, restlessness, and sleep disturbance when challenged with specific azo dyes like tartrazine at doses of 50mg+. A systematic review explicitly concludes that azo dye food colourants cause disturbed behaviour in children, and neurological toxicity mechanisms are well-documented.\\

- Reasons to Refute: The behavioral effects are not universal—only children identified as sensitive/reactors demonstrate responses—while average dietary intakes in the general population fall below the threshold doses that produce detectable effects. Effects appear to be modulated by individual susceptibility, genetic factors, and co-exposure to other additives rather than being a direct universal causation.\\

- Consensus \& Context Check: The claim's universal language ("cause") does not fully match the evidence, which shows a conditional association restricted to responsive children populations and at challenging doses above typical consumption levels. The relationship is best characterized as a risk factor for a subset of children rather than a universal cause for all cases of ADHD/hyperactivity.\\

- Final Justification: The accumulated evidence from meta-analyses and rigorous double-blind trials supports that azo food dyes can cause behavioral changes and hyperactivity-like symptoms in susceptible children, particularly at higher doses, but this is not a universal causal effect applicable to all children.\\

**Conclusion**\\
<answer>SUPPORTED</answer>
\end{promptblock}

\section{Broader Impacts}
The primary objective of \name{} is to mitigate the spread of health misinformation by verifying dubious claims. While our agent demonstrates promising capabilities, its accuracy in discerning all forms of misinformation is still not perfect. Therefore, we expect to use this agent as an assistive reference tool rather than a definitive authority. By providing users with comprehensive fact-checking reports, it is intended to aid human judgment. Over-reliance on our model for critical medical or health-related decision-making poses potential risks, and we strongly encourage users to consult professional medical advice.